\documentclass[sigconf,screen]{acmart}

\usepackage{xspace}
\usepackage{bm}
\usepackage{algorithm}
\usepackage{algorithmic}
\usepackage{placeins}
\usepackage{dsfont}
\usepackage{url}
\usepackage{svg}
\usepackage[export]{adjustbox}
\usepackage{subfig}
\usepackage{todonotes}
\usepackage{soul}
\usepackage{multirow}

\usepackage{amsmath,amssymb}
\usepackage{graphicx}

\DeclareMathOperator*{\TopK}{TopK}

\AtBeginDocument{%
  }

\copyrightyear{2026}
\acmYear{2026}
\setcopyright{cc}
\setcctype{by}
\acmConference[KDD 2026] {Proceedings of the 32nd ACM SIGKDD Conference on Knowledge Discovery and Data Mining V.2}{August 9--13, 2026}{Jeju Island, Republic of Korea.}
\acmBooktitle{Proceedings of the 32nd ACM SIGKDD Conference on Knowledge Discovery and Data Mining V.2 (KDD 2026), August 9--13, 2026, Jeju Island, Republic of Korea}
\acmISBN{979-8-4007-2259-2/2026/08}
\acmDOI{10.1145/3770855.3818945}

\newcommand{\junk}[1]{}
\newcommand{\method}{\textsc{EpiAwareNet}\xspace}
\newcommand{\methodnnpu}{\textsc{EpiAwareNet}\text{-}\textsc{nnPU}\xspace}

\begin{document}

\title{Prior-Guided Multi-Omic Transformers for Single-Cell Gene Regulatory Network Inference}





\author{Tianyang Xu}
\affiliation{
  \department{Elmore Family School of Electrical and Computer Engineering}
  \institution{Purdue University}
  \city{West Lafayette}
  \state{IN}
  \country{USA}
}
\email{xu1868@purdue.edu}

\author{Tianci Liu}
\affiliation{
  \department{Elmore Family School of Electrical and Computer Engineering}
  \institution{Purdue University}
  \city{West Lafayette}
  \state{IN}
  \country{USA}
}
\email{liu3351@purdue.edu}

\author{Niraj Rayamajhi}
\affiliation{
  \department{Department of Horticulture and Landscape Architecture}
  \institution{Purdue University}
  \city{West Lafayette}
  \state{IN}
  \country{USA}
}
\email{nrayamaj@purdue.edu}

\author{Ryan Patrick}
\affiliation{
  \department{School of Biological Sciences}
  \institution{Illinois State University}
  \city{Normal}
  \state{IL}
  \country{USA}
}
\email{rmpatr2@ilstu.edu}

\author{Kranthi Varala}
\affiliation{
  \department{Department of Horticulture and Landscape Architecture}
  \institution{Purdue University}
  \city{West Lafayette}
  \state{IN}
  \country{USA}
}
\email{kvarala@purdue.edu}

\author{Ying Li}
\affiliation{
  \department{Department of Horticulture and Landscape Architecture}
  \institution{Purdue University}
  \city{West Lafayette}
  \state{IN}
  \country{USA}
}
\email{li2627@purdue.edu}

\author{Jing Gao}
\affiliation{
  \department{Elmore Family School of Electrical and Computer Engineering}
  \institution{Purdue University}
  \city{West Lafayette}
  \state{IN}
  \country{USA}
}
\email{jinggao@purdue.edu}
\renewcommand{\shortauthors}{Tianyang et al.}
\def\authors{Tianyang Xu, Tianci Liu, Niraj Rayamajhi, Ryan Patrick, Kranthi Varala, Ying Li, Jing Gao}
\begin{abstract}
Gene regulatory networks (GRNs) capture transcription factor-target interactions and are central to understanding cell-state regulation and disease.
Reconstructing GRNs from paired single-cell transcriptomic and chromatin accessibility data is promising but challenging: scATAC is extremely sparse, and most methods rely on fixed peak-to-gene links and weak supervision.
We present \method, a prior-guided multi-omic Transformer framework that reconstructs GRNs from paired single-cell data using only lightweight biological priors.
In Stage 1, \method learns joint gene-peak representations with a gene--peak cross-attention module, enabling data-driven, gene-specific aggregation of accessibility signals rather than hard-coded peak-to-gene assignments. In Stage 2, \method incorporates a bulk-derived GRN prior as noisy positive edges to provide weak supervision under label scarcity, refining regulatory scores while remaining robust to prior noise.
In our experiments, \method improves overall GRN reconstruction robustness over representative single- and multi-omic baselines and yields GRNs with greater biological plausibility, such as improved recovery of known regulatory interactions,
suggesting that lightweight biological priors from bulk data can effectively guide single-cell GRN inference when combined with adaptive cross-modal representation learning.
Code and data are available at \url{https://github.com/tianyang-x/EpiAwareNet_pub}.

\end{abstract}



\maketitle

\section{Introduction}

In biology, gene regulatory networks (GRNs) describe how regulators, such as transcription factors (TFs), control downstream target genes and thereby shape cellular states.
Characterizing regulator--target relationships is fundamental for understanding developmental patterning and morphogenesis~\cite{Davidson2006Book}, cell-type specification and differentiation~\cite{Karlebach2008}, and responses to environmental stimuli~\cite{Faith2007}. GRN inference is also central to applied settings, including drug-target discovery~\cite{DeSmet2010}, synthetic biology and cell reprogramming~\cite{Davidson2006Book}, and agricultural or industrial biotechnology~\cite{Karlebach2008}.
From a machine learning perspective, GRN inference requires learning directed regulatory relationships from observational transcriptomic data that are often noisy, sparse, and only partially informative.
Consequently, reconstructing accurate GRNs is a core task at the intersection of computational biology and machine learning.

Early \emph{in silico} efforts inferred GRNs from bulk transcriptomic data, in which many cells are pooled and each gene is represented by an average expression level over a mixed population, using heuristic or regression-based approaches such as ARACNe and GENIE3~\cite{Margolin2006,HuynhThu2010}.
Because bulk measurements capture only coarse population-level patterns, their predictions are limited in resolving cell-type- or condition-specific regulatory edges.
With the advent of single-cell technologies~\cite{Macosko2015,Zheng2017}, GRN reconstruction expanded to single-cell RNA sequencing (scRNA-seq), enabling cell-type-specific and context-dependent regulatory analysis~\cite{Trapnell2014,PijuanSala2019}.
However, extreme sparsity and noise in scRNA-seq make it difficult to distinguish genuine regulatory signals from stochastic fluctuations~\cite{Kharchenko2014}, often yielding unstable networks~\cite{Pratapa2020,Chen2018SCGRN} or predictions with weak support from independent evidence~\cite{Akers2021Review}.

A natural way to mitigate these limitations is to integrate scRNA-seq with an additional omic layer measured in the same cells (a ``multi-omic'' setting), such as single-cell ATAC sequencing (scATAC-seq), which measures chromatin accessibility, i.e., whether local DNA regions are ``open'' and thus more likely to support regulatory activity.
Intuitively, two noisy views can still be beneficial when they provide independent constraints on the underlying regulatory relationships.
When multiple molecular modalities (e.g., RNA expression and chromatin accessibility) are profiled in the same cells, the resulting paired measurements are commonly referred to as a ``multi-omic'' dataset.
In principle, accessibility marks candidate regulatory regions (e.g., promoters and enhancers) and provides complementary evidence that can support weak or noisy expression measurements~\cite{Stuart2019,Hao2021}.
Yet multi-omic integration introduces two key challenges for GRN inference. First, per-cell chromatin accessibility is even sparser and noisier than expression, and linking accessibility peaks to their target genes is fundamentally ambiguous. Second, enhancer--promoter regulation is context-dependent~\cite{Schoenfelder2019LongRange}; therefore, fixed peak-to-gene heuristics can assign accessibility evidence to incorrect targets, limiting robustness across cell types or conditions. Existing solutions often rely on heuristic peak-to-gene linkages, for example by associating TF motif accessibility with gene expression (e.g., SCENIC+) or by inferring co-accessibility between enhancers and promoters (e.g., Cicero)~\cite{GonzalezBlas2023,Pliner2018}. However, such heuristics can be brittle when regulatory interactions deviate from the assumed linking rules~\cite{Schoenfelder2019LongRange}.
Overall, robust multi-omic GRN inference remains challenging because it requires assigning sparse cross-modal evidence to correct targets without reliable peak-to-gene ground truth.
Recently, foundation models pretrained on large-scale single-cell datasets have shown promise for learning transferable cellular representations~\cite{Theodoris2023,Cui2024}.
However, it remains unclear whether such representations, when adapted to GRN reconstruction, yield regulator--target predictions that are both accurate and supported by independent regulatory evidence in sparse multi-omic settings.
To address this gap, we evaluate inferred GRNs not only by predictive performance (i.e., how many reference edges are predicted), but also by orthogonal biological evidence through literature-grounded case studies on representative regulators.

These challenges motivate a prior-guided formulation: we aim to recover regulator--target edges from sparse multi-omic observations over a large combinatorial space with limited reliable labels, while avoiding hard-coded peak-to-gene mappings and leveraging weak biological supervision under label scarcity.
Our work is therefore situated in the broader prior-informed GRN literature, where external information such as motifs, genomic proximity, chromatin accessibility, and known regulatory links is used to constrain otherwise underdetermined network inference problems~\cite{Stock2025Prior}.
Rather than claiming that each component is new in isolation, our contribution is the specific formulation for paired scRNA-seq/scATAC-seq GRN refinement: paired multi-omic input, candidate-constrained gene--peak cross-attention for adaptive peak aggregation, and a separate weakly supervised regulator--target scoring stage that uses bulk-derived edges as noisy positives rather than hard constraints.

Here, we present \method, a prior-guided multi-omic Transformer framework that integrates weak candidate priors with deep representation learning for GRN reconstruction from paired scRNA-seq and scATAC-seq data. Specifically, \method follows a two-stage design.

In Stage~1 (\textbf{representation learning stage}), \method jointly models gene expression and chromatin accessibility with a Transformer-style architecture. Rather than fixing peak-to-gene mappings, \method constructs a candidate set of ATAC peaks for each gene using genomic proximity as a lightweight prior, and uses sparse cross-attention to aggregate accessibility signals when forming each gene representation. This masked-reconstruction stage is a proxy for learning chromatin-informed representations: it can capture how accessible candidate regions help explain expression variation, but it does not by itself identify causal TF--target regulation or edge direction.

In Stage~2 (\textbf{GRN fine-tuning stage}), \method is adapted for directional GRN reconstruction by attaching a lightweight prediction head and incorporating a bulk-derived GRN as a weak biological prior. Directionality is introduced at this stage by scoring ordered regulator--target pairs under directed weak supervision. Concretely, edges in the bulk GRN are treated as noisy positives that provide regulatory cues, while remaining regulator--target pairs are left unlabeled. The bulk GRN is \emph{not} treated as ground truth to be reproduced; instead, it acts as a noisy edge-level guide and regularizer over the paired RNA--ATAC representations learned in Stage~1.
Because the bulk GRN provides only noisy positives while the rest of regulator--target pairs remain unlabeled, Stage~2 naturally falls into a positive--unlabeled (PU) setting.
\method uses a simple binary cross-entropy (BCE) objective, in which unlabeled regulator--target pairs are treated as negatives during training, thereby ``pushing down'' most unlabeled pairs and producing a well-separated score ordering over the full candidate space.
We additionally evaluate a variant of \method, \methodnnpu, which explicitly accounts for the possibility of hidden positives among unlabeled pairs and is therefore more conservative in penalizing the unlabeled set; in practice, this often concentrates probability mass on a smaller set of high-confidence edges, benefiting precision among the very top-ranked predictions under small screening budgets while providing less fine-grained separation among the remaining edges.

Our evaluation combines established paired multi-omic benchmarks (PBMC and mouse brain) with the tomato \texttt{pN}/\texttt{mN} root multiome dataset~\cite{Patrick2026Every}, extending the assessment to a plant setting and supporting a case study on regulators unseen during training.

In summary, our contributions are threefold.
\textbf{(1) Prior-guided cross-modal representation learning:} we formulate paired scRNA-seq/scATAC-seq GRN inference as a prior-guided multi-omic representation learning problem, using sparse gene--peak cross-attention to learn accessibility evidence within weak genomic candidate sets rather than relying on fixed peak-to-gene mapping rules.
\textbf{(2) Weakly supervised GRN refinement with noisy priors:} we use bulk-derived GRN edges as noisy positives while leaving remaining regulator--target pairs unlabeled, and analyze BCE vs.\ nnPU objectives as a trade-off between global ranking and very-top precision.
\textbf{(3) Comprehensive evaluation with predictive and biological validity:} we evaluate paired multi-omic datasets across mammalian and plant systems, showing improved overall GRN reconstruction and biologically plausible case-study predictions supported by prior literature.
\section{Background and Related Works}

\paragraph{From bulk GRNs to multi-omic single-cell inference.}
Classical \emph{in silico} GRN inference from bulk expression, i.e., measurements obtained by pooling multiple cells and tissues. They typically rely on statistical dependency or regression-based modeling (e.g., correlation/MI or tree-based regressors)~\cite{Langfelder2008,Faith2007,HuynhThu2010,Moerman2019}.
While the methods are scalable, bulk profiles average over heterogeneous cell populations and obscure context-specific regulation.
Single-cell RNA sequencing (scRNA-seq) quantifies transcriptomes at the level of \textbf{individual cells}. It enables GRN inference at cellular resolution but is challenged by severe sparsity and technical noise~\cite{Kharchenko2014}.
Recent paired scRNA-seq/scATAC-seq approaches incorporate chromatin accessibility to provide regulatory evidence, often via motif deviation scores or co-accessibility graphs~\cite{Schep2017,Pliner2018}.
However, many pipelines still depend on heuristic peak-to-gene assignment (e.g., distance thresholds) as a fixed preprocessing step, which can miss distal enhancer--gene links and fail to capture cell-type/condition-specific enhancer--promoter interactions.
In contrast, \method uses genomic proximity and simple statistics only to define candidate peak sets per gene, and then learns candidate-constrained gene--peak attention \emph{within} each set in a data-driven manner.

\paragraph{Task-specific modeling and weak supervision with priors.}
Transformer-based foundation models (e.g., Geneformer, scGPT) learn transferable gene/cell embeddings for downstream tasks, but are not optimized for GRN edge discovery and do not explicitly model peak-level regulatory evidence~\cite{Theodoris2023,Cui2024}.
Recent reviews emphasize that prior information is often essential for GRN inference from sparse single-cell data, including TF motif annotations, genomic proximity, chromatin accessibility, perturbation data, curated interaction databases, and pathway constraints~\cite{Stock2025Prior}.
Within this prior-informed landscape, methods differ in whether priors are used as fixed preprocessing constraints, graph structures, or weak labels.
\method instead targets paired multi-omic GRN refinement by coupling candidate-constrained gene--peak attention with weakly supervised regulator--target scoring.
This formulation uses the bulk-derived prior edges as a soft guide rather than a hard constraint, so the model can reduce the influence of prior edges that are not supported by the single-cell multi-omic evidence.
\section{Methods}

\begin{figure*}[t]
    \centering
    \includegraphics[width=0.9\linewidth]{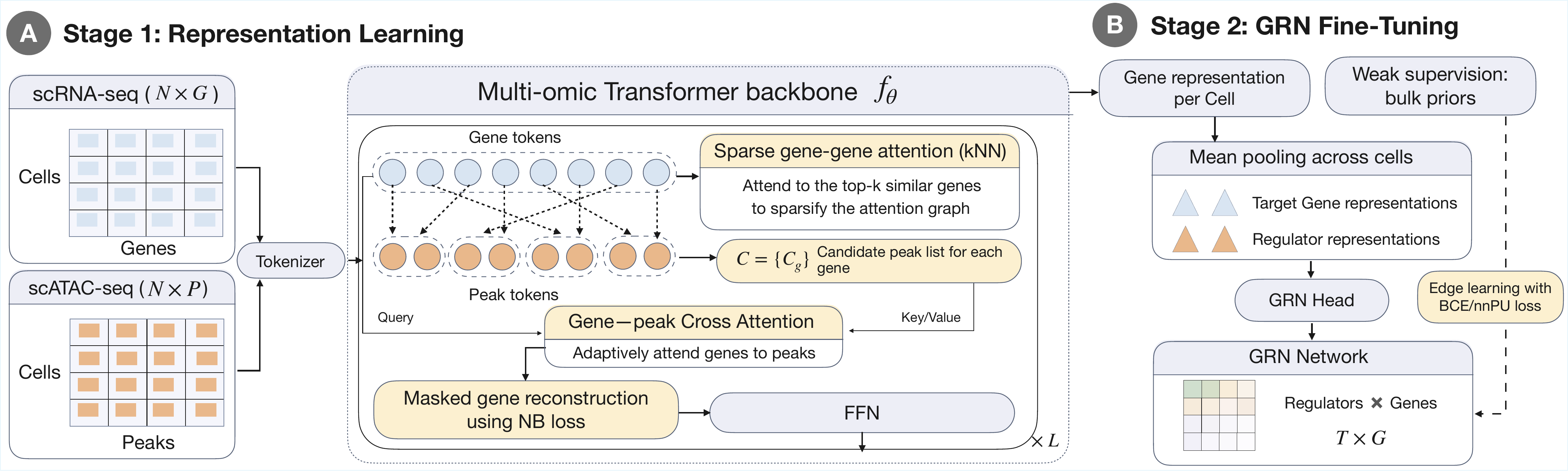}
    \caption{Overview of the EpiAwareNet framework. Stage~1 encodes paired scRNA-seq/scATAC-seq with a prior-guided multi-omic Transformer, using candidate-constrained gene-centric cross-attention to integrate chromatin context. Stage~2 aggregates gene representations and scores regulator--target edges with weak supervision from a bulk-derived prior.}
    \label{fig:main}
    \vspace{-7pt}
\end{figure*}

We formalize single-cell multi-omic GRN inference as a prior-guided learning problem and introduce \method (Figure~\ref{fig:main}), which combines a multi-omic Transformer backbone with two lightweight priors:
(i) a gene--peak candidate set $\mathcal{C}=\{\mathcal{C}_g\}$ that restricts gene--peak interactions to a small, plausible neighborhood per gene, and
(ii) a bulk-derived positive regulatory links $\mathcal{P}_{\mathrm{bulk}}$ that serves as noisy positive supervision.
Implementation details are provided in Appendix~\ref{app:impl_details}.

\subsection{Problem Formulation}

Let $\mathcal{X}_{gene} \in \mathbb{R}^{N \times G}$ be the gene expression matrix for $N$ cells and $G$ genes, and let $\mathcal{X}_{peak} \in \mathbb{R}^{N \times P}$ be the chromatin accessibility matrix for $P$ peaks, where each peak corresponds to a genomic region with open chromatin measured by scATAC-seq.
We denote the index sets of genes and peaks by $\mathcal{G}$ and $\mathcal{P}$, respectively.
Let $\mathcal{T} \subseteq \mathcal{G}$ be the index set of known regulators, including both transcription factors and chromatin regulators, and let $T = |\mathcal{T}|$.
For brevity, we refer to elements in $\mathcal{T}$ as \emph{regulators}.
Our goal is to infer a gene regulatory network (GRN) represented as a weighted adjacency matrix $\mathcal{A} \in [0,1]^{T \times G}$, where each entry $\mathcal{A}_{t,g}$ is the confidence score of a directed regulatory interaction $t \rightarrow g$.


Inferring $\mathcal{A}$ from single-cell data is challenging because both $\mathcal{X}_{gene}$ and $\mathcal{X}_{peak}$ are high-dimensional and sparse at the per-cell level.
A primary difficulty lies in linking peaks to genes: without constraints, there are $O(GP)$ potential gene--peak pairs to consider.
To guide search over this large space, \method incorporates two priors:

\begin{enumerate}
    \item \textbf{Gene--peak candidate sets $\mathcal{C}$.}
    For each gene $g \in \mathcal{G}$, we define a candidate set of peaks $\mathcal{C}_g \subseteq \mathcal{P}$ with $|\mathcal{C}_g| \ll P$.
    In our implementation, $\mathcal{C}_g$ is provided as a precomputed gene--peak candidate list of fixed maximum length (Appendix~\ref{app:data_prep}).
    This prior restricts cross-modal aggregation to a small subset of peaks per gene.

    \item \textbf{Bulk-derived positive regulatory links $\mathcal{P}_{\text{bulk}}$.}
    We use a set of bulk-derived regulator--target gene links $\mathcal{P}_{\text{bulk}} \subseteq \mathcal{T}\times\mathcal{G}$ as noisy positives.
    All remaining regulator--target gene pairs are treated as unlabeled candidates.
\end{enumerate}

Details on the construction of these priors from real data are provided in Appendix~\ref{app:data_prep}.
Our \method framework consists of a two-stage model training.
In \textbf{Stage 1}, we train a Transformer-based backbone $f_\theta$ to learn peak-aware gene representations from both $\mathcal{X}_{gene}$ and $\mathcal{X}_{peak}$, 
using $\mathcal{C}$ to constrain cross-modal attention.
In \textbf{Stage 2}, we freeze $f_\theta$, attach a lightweight prediction head, and train it using $\mathcal{P}_{\text{bulk}}$ with sampled unlabeled pairs to infer the GRN.
We now detail the two stages.

\subsection{Stage 1: Representation Learning}

Stage~1 trains a backbone $f_\theta$ that maps multiomic features (i.e. $\mathcal{X}_{gene}$ and $\mathcal{X}_{peak}$) to cell-specific latent representations of genes and peaks.
For each cell $n \in \{1,\dots,N\}$, the backbone produces gene and peak representations
\[
H^{(L)}_{gene,n},\; H^{(L)}_{peak,n}
=
f_\theta\big(\mathbf{x}^{RNA}_n, \mathbf{x}^{ATAC}_n \,\big|\, \mathcal{C}\big),
\]
where
\[
H^{(L)}_{gene,n} = [\,\mathbf{h}^{(L)}_{n,g}\,]_{g \in \mathcal{G}},
\qquad
H^{(L)}_{peak,n} = [\,\mathbf{h}^{(L)}_{n,p}\,]_{p \in \mathcal{P}}.
\]
From now on, we omit the cell index $n$ for brevity if it's clear from the context.

\subsubsection{Embedding Layer}

For a given cell $n$, let $x^{gene}_{n,g}\in\mathbb{R}$ denote the preprocessed expression of gene feature $g$, and let $x^{peak}_{n,p}\in\mathbb{R}$ denote the preprocessed accessibility of peak feature $p$.

We project these scalar measurements through separate linear layers:
\[
\mathbf{e}^{(0)}_{n,g} = W_{RNA} \, x^{RNA}_{n,g} + \mathbf{b}_{RNA}, \quad
\mathbf{e}^{(0)}_{n,p} = W_{ATAC} \, x^{ATAC}_{n,p} + \mathbf{b}_{ATAC},
\]
and add type embeddings:
\[
\mathbf{h}^{(0)}_{n,g} = \mathbf{e}^{(0)}_{n,g} + \mathbf{s}_{gene}, \quad
\mathbf{h}^{(0)}_{n,p} = \mathbf{e}^{(0)}_{n,p} + \mathbf{s}_{peak}.
\]

\subsubsection{Gene--Gene and Gene--Peak Attention}

The backbone consists of $L$ Transformer blocks.
Each block contains (i) sparse gene--gene self-attention, (ii) candidate-constrained gene--peak cross-attention applied to gene tokens, and (iii) a feed-forward update for peak tokens.
Each sub-layer is wrapped by residual connections and layer normalization.

\paragraph{Gene--gene sparse self-attention.}
Full self-attention over $G$ genes costs $O(G^2)$ in time complexity.
To sparsify the attention graph and lower the memory cost, we restrict each gene to attend only to its $k_{\text{NN}}$ nearest neighbors in the current latent space.
At layer $l$, for each gene token $g\in\mathcal{G}$, we compute cosine similarity between $\mathbf{h}^{(l)}_g$ and all gene representations $\{\mathbf{h}^{(l)}_{g'}\}_{g'\in\mathcal{G}}$, and define
\[
\mathcal{N}^{(l)}_g
=
\TopK_{g'\in\mathcal{G}}\,
\cos\!\big(\mathbf{h}^{(l)}_g,\mathbf{h}^{(l)}_{g'}\big),
\qquad
|\mathcal{N}^{(l)}_g| = k_{\text{NN}}.
\]
We then apply standard multi-head self-attention where the query is $\mathbf{h}^{(l)}_g$ and keys/values are restricted to $\{\mathbf{h}^{(l)}_{g'}\}_{g'\in\mathcal{N}^{(l)}_g}$.
In practice, we reuse the same neighbor indices across Transformer blocks for efficiency (Appendix~\ref{app:impl_details}).

\paragraph{Gene--peak candidate-constrained cross-attention.}
To inject chromatin context, each gene $g$ attends only to peaks in its candidate set $\mathcal{C}_g$.
Within block $l$, we apply multi-head cross-attention, in which the gene representation after gene--gene attention is used as the query.
Concretely, let $\widetilde{\mathbf{h}}^{(l)}_g$ denote the gene embedding after the gene--gene attention sub-layer (including its residual and normalization), and let $\mathbf{h}^{(l)}_p$ denote the peak embedding.
We compute scaled dot-product attention logits over candidate peaks:
\[
s^{(l)}_{g,p}
=
\frac{\langle W_Q \widetilde{\mathbf{h}}^{(l)}_g,\; W_K \mathbf{h}^{(l)}_p\rangle}{\sqrt{d}},
\qquad p\in\mathcal{C}_g,
\]
where $W_Q,W_K$ are head-specific projections (omitted for brevity when multi-head is clear).
Optionally, for efficiency we retain only the Top-$K$ candidate peaks according to these logits,
\[
\mathcal{I}^{(l)}_{\text{topK}}(g)
=
\TopK_{p\in\mathcal{C}_g}\, s^{(l)}_{g,p},
\qquad
|\mathcal{I}^{(l)}_{\text{topK}}(g)|=K,
\]
and apply a masked (renormalized) softmax over the retained subset:
\[
\alpha^{(l)}_{g,p}
=
\frac{\exp(s^{(l)}_{g,p})}{\sum_{p'\in\mathcal{I}^{(l)}_{\text{topK}}(g)} \exp(s^{(l)}_{g,p'})},
\qquad p\in\mathcal{I}^{(l)}_{\text{topK}}(g),
\]
with $\alpha^{(l)}_{g,p}=0$ for $p\notin\mathcal{I}^{(l)}_{\text{topK}}(g)$.
The cross-attention output aggregated peak features from the selected candidates:
\[
\text{Out}^{(l)}_{\text{GP}}(g)
=
\sum_{p \in \mathcal{I}^{(l)}_{\text{topK}}(g)}
\alpha^{(l)}_{g,p}\, (W_V \mathbf{h}^{(l)}_p),
\]
where $W_V$ is the value projection (head-specific in the multi-head case).
This design makes the gene$\rightarrow$peak interaction cost scale with $\sum_g |\mathcal{C}_g|$ (and with $\sum_g K$ when Top-$K$ is enabled), rather than the full number of peaks $P$.
Importantly, $\mathcal{C}_g$ only defines a weak candidate neighborhood; the model still performs data-driven, soft selection within $\mathcal{C}_g$ via attention weights, instead of imposing a fixed peak-to-gene mapping.

\paragraph{Fusion and propagation.}
Let $\text{Out}^{(l)}_{\text{GP}}(g)$ denote the gene-side cross-attention output.
We update gene tokens by a residual + normalization step that fuses $\text{Out}^{(l)}_{\text{GP}}(g)$ into $\mathbf{h}^{(l)}_g$.
Peak tokens do not attend back to genes; they are propagated by a lightweight feed-forward update, keeping cross-modal coupling sparse and gene-centric.

\subsubsection{Pretraining Objective}
We pretrain $f_\theta$ using a masked reconstruction objective over gene expression.
For each cell $n$, we randomly mask a subset $\mathcal{M}_n \subseteq \mathcal{G}$ of genes (15\% by default) and reconstruct their expression values from the final gene representations.
We use a negative-binomial-style (NB) reconstruction loss, 
parameterizing the NB mean and dispersion from $\mathbf{h}^{(L)}_{n,g}$ via a small decoder network.
Here we use a negative binomial (NB) reconstruction loss because scRNA-seq expression is typically noisy and variable, and NB is a simple way to model this behavior better than a Poisson or Gaussian loss.
\[
\mathcal{L}_{\text{MLL}}
=
- \sum_{n=1}^{N}
  \sum_{g \in \mathcal{M}_n}
  \log \mathcal{P}_{NB}\!\left(
      x^{gene}_{n,g}
      \,\middle|\,
      \mathbf{h}^{(L)}_{n,g}; \theta
  \right).
\]
We apply this loss to the preprocessed expression values used by the model (Appendix~\ref{app:impl_details}), and treat it as a practical reconstruction objective that stabilizes training under overdispersed single-cell signals.

\paragraph{Design rationale.}
Candidate-constrained cross-attention reduces computation from $O(NGPd)$ to approximately $O\!\left(N\sum_g |\mathcal{C}_g|d\right)$ by restricting gene--peak aggregation to neighborhoods $\mathcal{C}_g$ (Appendix~\ref{app:complexity}).

\subsection{Stage 2: GRN Fine-Tuning}

In Stage~2 we introduce the directed GRN prediction task. To this end, we freeze the backbone $f_\theta$ and learn a lightweight prediction head $h_\phi$ on top of its representations to infer ordered regulator--target gene edges from bulk-derived directed priors in a positive--unlabeled setting.
Thus, Stage~1 provides chromatin-informed representations, whereas Stage~2 supplies regulator$\rightarrow$target directionality and edge-level weak supervision.
The bulk-derived prior is used as a noisy guide and regularizer rather than the sole source of signal; Section~\ref{sec:robustness} further examines how performance changes as prior edges are dropped.

\subsubsection{GRN Prediction Head}

For a given training collection of cells (e.g., all cells in the dataset), we obtain aggregated representations for each regulator and gene by pooling their cell-specific embeddings from Stage~1.
Let $\mathbf{h}^{(L)}_{n,t}$ and $\mathbf{h}^{(L)}_{n,g}$ denote the final-layer representations of regulators $t \in \mathcal{T}$ and gene $g \in \mathcal{G}$ in cell $n$.
We form context-level embeddings by simple averaging over cells,
\[
\bar{\mathbf{h}}^{(L)}_{t} = \frac{1}{N} \sum_{n=1}^{N} \mathbf{h}^{(L)}_{n,t},
\qquad
\bar{\mathbf{h}}^{(L)}_{g} = \frac{1}{N} \sum_{n=1}^{N} \mathbf{h}^{(L)}_{n,g},
\]
and write
\[
\mathbf{z}_{t} = \bar{\mathbf{h}}^{(L)}_{t}, \qquad
\mathbf{z}_{g} = \bar{\mathbf{h}}^{(L)}_{g}
\]
for brevity.
Because the head scores pairs after this averaging step, the inferred network should be interpreted as a context-level (cell-set-level) GRN for the input collection of cells, rather than a distinct GRN for each individual cell.
The prediction head $h_\phi$ takes the concatenated representation of a regulator--target gene pair and outputs an unnormalized score
\[
s_{t,g} = h_\phi\big([\mathbf{z}_{t} ; \mathbf{z}_{g}]\big),
\quad (t,g) \in \mathcal{T} \times \mathcal{G},
\]
which is converted to a probability via sigmoid function $\sigma(\cdot)$:
\[
\hat{a}_{t,g} = \sigma(s_{t,g}) = (\exp(-s_{t,g}) + 1)^{-1}  \in (0,1).
\]
In our implementation, $h_\phi$ uses a shallow MLP layer, making Stage~2 computationally lightweight.
During fine-tuning, only $\phi$ is updated while the backbone parameters $\theta$ remain frozen.

\subsubsection{Weak-supervision objective}

Let $\mathcal{P}_{\text{bulk}} \subseteq \mathcal{T}\times\mathcal{G}$ denote bulk-derived regulator--target gene pairs used as noisy positives, and let
$\mathcal{U} = (\mathcal{T}\times\mathcal{G}) \setminus \mathcal{P}_{\text{bulk}}$
denote the remaining unlabeled pairs.
Since enumerating $\mathcal{U}$ is infeasible, we approximate expectations with minibatches consisting of positives sampled from $\mathcal{P}_{\text{bulk}}$ and unlabeled pairs sampled from $\mathcal{U}$ (Appendix~\ref{app:impl_details}).

\paragraph{\method objective (BCE with sampled unlabeled pairs).}
\method treats sampled unlabeled pairs as negatives during training and uses a binary cross-entropy (BCE) loss:
\[
\mathcal{L}_{\text{BCE}}
=
\mathbb{E}_{(t,g)\sim \mathcal{P}_{\text{bulk}}}\big[\ell_{\text{BCE}}(1, s_{t,g})\big]
+
\mathbb{E}_{(t,g)\sim \mathcal{U}_b}\big[\ell_{\text{BCE}}(0, s_{t,g})\big],
\]
where $\mathcal{U}_b$ denotes unlabeled pairs sampled in the minibatch and
$\ell_{\text{BCE}}(y,s) = -y\log\sigma(s)-(1-y)\log(1-\sigma(s))$.
Unless otherwise stated, we use a fixed positive:unlabeled sampling ratio during training (Appendix~\ref{app:impl_details}).

\paragraph{Alternative objective (nnPU) for \methodnnpu}
For \methodnnpu, we also implement the non-negative positive--unlabeled (nnPU) objective~\cite{kiryo2017nnpu}. 
With a margin loss $\ell(y,s)=\max(0,1-y\cdot s)$ and positive class prior $\pi_P$, we define
\begin{align*}
\widehat{R}^{+}(\phi)
&= \pi_P \, \mathbb{E}_{(t,g)\sim \mathcal{P}_{\text{bulk}}}\big[\ell(+1, s_{t,g})\big],\\
\widehat{R}^{-}(\phi)
&= \mathbb{E}_{(t,g)\sim \mathcal{U}_b}\big[\ell(-1, s_{t,g})\big]
   - \pi_P \, \mathbb{E}_{(t,g)\sim \mathcal{P}_{\text{bulk}}}\big[\ell(-1, s_{t,g})\big],
\end{align*}
and optimize
\[
\mathcal{L}_{\text{nnPU}}(\phi) = \widehat{R}^{+}(\phi) + \max\big(0, \widehat{R}^{-}(\phi)\big).
\]
In our implementation, $\pi_P$ is treated as a hyperparameter (Appendix~\ref{app:impl_details}).

\subsubsection{GRN Extraction.}

Once $h_\phi$ is trained, the composed model $h_\phi \circ f_\theta$ can be applied to infer GRNs for cells $C$.
Given single-cell data for cells $c \in C$, we first fetch embeddings by averaging representations over $C$:
\[
\bar{\mathbf{h}}^{(L),C}_{t}
=
\frac{1}{|C|}
\sum_{c \in C}
\mathbf{h}^{(L)}_{c,t},
\qquad
\bar{\mathbf{h}}^{(L),C}_{g}
=
\frac{1}{|C|}
\sum_{c \in C}
\mathbf{h}^{(L)}_{c,g},
\]
and then apply the prediction head to these aggregated embeddings:
\[
s^{C}_{t,g}
=
h_\phi\big([\bar{\mathbf{h}}^{(L),C}_{t} ; \bar{\mathbf{h}}^{(L),C}_{g}]\big),
\qquad
\mathcal{A}^C_{t,g}
=
\sigma\big(s^{C}_{t,g}\big),
\quad
t \in \mathcal{T}, \; g \in \mathcal{G}.
\]
Each $\mathcal{A}^C_{t,g}$ thus represents the confidence that regulator $t$ regulates gene $g$ within the cell set $C$.
\section{Evaluation Results}
\label{sec:eval}

\subsection{Experiment Setup}
\label{sec:eval_setup}

\subsubsection{Evaluation datasets.}
\begin{table}[t]
\centering
\footnotesize
\caption{
Summary of datasets.
$N$, $G$, and $P$ denote the number of cells, genes, and peaks, $|\mathcal{E}|$ the size of the regulator--target edges, and $|\mathcal{T}|$ the size of regulators.
}
\label{tab:datasets}
\begin{tabular}{l l r r r r r}
\toprule
Dataset & Species & $N$ & $G$ & $P$ & $|\mathcal{E}|$ & $|\mathcal{T}|$ \\
\midrule
mN & tomato & 8{,}748 & 34{,}074 & 53{,}570 & 170{,}300 & 2{,}381 \\
pN & tomato & 12{,}448 & 34{,}074 & 55{,}930 & 170{,}300 & 2{,}381 \\
PBMC & human & 12{,}012 & 36{,}600 & 111{,}856 & 5{,}092 & 118 \\
mouse brain & mouse & 4{,}881 & 32{,}284 & 42{,}755 & 5{,}064 & 819 \\
\bottomrule
\end{tabular}
\vspace{-2em}
\end{table}

We evaluate \method on four paired multiome datasets spanning plant and mammalian systems, where RNA and ATAC profiles are measured for the same cells.
\begin{itemize}
    \item \textbf{Tomato (\texttt{pN} and \texttt{mN}).}
    This is a tomato root multiome dataset under normal (\texttt{pN}) and low-nitrogen (\texttt{mN}) conditions~\cite{Patrick2026Every}.
    For Stage~2, we use a dataset-specific regulator list and bulk-derived regulator--target links provided with the tomato benchmark.

    \item \textbf{PBMC.}
    This is a human PBMC multiome dataset from 10x Genomics\footnote{\url{https://www.10xgenomics.com/datasets/10-k-human-pbm-cs-multiome-v-1-0-chromium-controller-1-standard-2-0-0}}, with matched RNA and ATAC measured from nuclei from a healthy donor.
    For Stage~2 weak supervision, we use a curated regulator list and regulator--target links from DoRothEA~\cite{GarciaAlonso2019DoRothEA}.

    \item \textbf{Mouse brain.}
    This is an embryonic mouse brain multiome dataset from 10x Genomics\footnote{\url{https://www.10xgenomics.com/datasets/fresh-embryonic-e-18-mouse-brain-5-k-1-standard-2-0-0}}.
    For Stage~2 weak supervision, we use a curated regulator list and regulator--target links from TRRUST~\cite{Han2018TRRUSTv2}.
\end{itemize}

For each dataset, we randomly split the curated regulator--target edges into training/validation (\texttt{train}/\texttt{val}) (80\%/20\%).
Only \texttt{train} edges are used for any supervised fitting (including Stage~2 and matched-head baselines), and we report metrics on held-out \texttt{val} edges.
To avoid evaluation leakage, we do not use \texttt{val} edges for checkpoint selection or hyperparameter tuning; unless otherwise stated, we report the final checkpoint under a fixed training schedule.
All methods use the same split.
Additional preprocessing details are provided in Appendix~\ref{app:data_prep}, and dataset statistics are summarized in Table~\ref{tab:datasets}.
Thus, this benchmark should be interpreted as weakly supervised generalization to unseen edges under a prior-informed regime, rather than prior-free de novo GRN discovery.

\subsubsection{Baselines.}
We compare \method against representative GRN inference baselines from three families: \textbf{classical RNA-only} methods (GRNBoost2 and WGCNA), \textbf{deep representation} baselines (scGPT and scGLUE with matched heads), and \textbf{recent paired multi-omic GRN pipelines} (SCENIC+ and Pando).
\paragraph{GRNBoost2 (RNA-only).}
GRNBoost2 fits a gradient-boosted tree model for each target gene and ranks candidate regulators using feature-importance scores~\cite{Moerman2019}.
We run it on the scRNA-seq matrix and restrict candidate regulators to the dataset regulator list $\mathcal{T}$.

\paragraph{WGCNA (RNA-only).}
WGCNA builds a weighted gene co-expression network via correlation-based similarity with soft-thresholding and module structure~\cite{Langfelder2008}.
Since WGCNA is undirected, we derive regulator$\rightarrow$gene scores by exporting regulator-anchored weighted links from the co-expression network (Appendix~\ref{app:baselines_impl}).

\paragraph{scGPT + matched head (RNA-only).}
scGPT is a Transformer trained on single-cell expression data to learn gene/cell representations~\cite{Cui2024}. Matched-head means a regulator--target prediction head with the same structure as \method.

\paragraph{scGLUE + matched head (RNA+ATAC)}
scGLUE learns aligned representations across modalities under a guidance graph linking features such as peaks and genes~\cite{Cao2022GLUE}.

To isolate representation quality from downstream scoring choices, we extract gene embeddings from scGPT and scGLUE, then train a regulator--target prediction head with the \emph{same architecture} and \emph{hyperparameters} as \method Stage~2 from scratch on top of these embeddings respectively, using the same regulator list and the same \texttt{train} prior edges.

\paragraph{SCENIC+ and Pando (RNA+ATAC).}
SCENIC+ infers enhancer-driven GRNs from paired gene expression and chromatin accessibility by combining region-to-gene links, motif enrichment, and tree-based regulatory modeling~\cite{GonzalezBlas2023}.
Pando models gene expression through TF--peak interactions from multimodal single-cell measurements~\cite{Fleck2023Pando}.
For fair comparison, both methods are evaluated on the same fixed candidate edge space $E=\mathcal{T}\times\mathcal{G}$ as other baselines; because these sparse-output pipelines return only selected edges, unreturned regulator--target pairs are assigned score 0.
Implementation details for all baselines are provided in Appendix~\ref{app:baselines_impl}.

\subsubsection{Evaluation metrics.}
We evaluate GRN prediction as ranking directed regulator$\rightarrow$gene edges over the shared candidate space $E=\mathcal{T}\times\mathcal{G}$.
We treat any unreported edge as the lowest-confidence prediction so that all methods, including sparse-output pipelines such as SCENIC+ and Pando, still induce a complete ranking over $E$.
We report AUPRC/AUROC for global ranking quality, and early-retrieval metrics at small screening budgets, namely in the top $K$ edges predicted (Precision@K and Hit@K; main table reports P@100).
AUPRC and AUROC summarize the area under the precision--recall curve and ROC curve over the full edge ranking, respectively.
To contextualize AUPRC under extreme class imbalance, we also report the AUPRC ratio, defined as AUPRC divided by the positive ratio $|E_{\mathrm{val}}|/|E|$ (expected AUPRC of a random ranking).
Additionally, we report Precision@K and Hit@K, which are the Precision and absolute number of correctly predicted edges of the top-K results retrieved by the model. They evaluate retrieval performance under the $K$ top-ranked predicted edges, which we call ``early-retrieval'' later.
Unless otherwise stated, all metrics are computed on the held-out \texttt{val} regulatory links.

\subsubsection{Implementation details.}
For each dataset, we train \method from scratch.
Stage~1 trains the multi-omic Transformer backbone on paired RNA/ATAC matrices after standard preprocessing (library-size normalization followed by log1p for RNA, i.e., normalizing raw counts $p$ as $\log (1+p)$; dataset-specific ATAC normalization with log-transform applied in the dataloader). 
Stage~1 uses candidate-constrained gene--peak cross-attention with precomputed $\pm 5$kb candidate peak sets to minimalize noise and candidate sizes.
Stage~2 freezes the backbone and trains only the regulator--target prediction head with a BCE objective, treating bulk-derived \texttt{train} edges as positives and uniformly sampled regulator--target pairs (excluding positives) as negatives (negative-to-positive ratio 10:1).
Full hyperparameters and reproducibility details are provided in Appendix~\ref{app:impl_details}.

\subsection{Overall Performance}
\label{sec:overall_grn_accuracy}

\begin{table*}[t]
\centering
\setlength{\tabcolsep}{3pt}
\renewcommand{\arraystretch}{1.05}
\scriptsize
\caption{
We report AUPRC, AUROC, Precision@100 (P@100), and AUPRC ratio (AUPRC divided by the positive-edge ratio; random ranking has expected ratio 1).
Best within each dataset are in \textbf{bold}. $^{*}$ indicates P@100 computed from canonical scores with tied unreturned edges assigned the lowest confidence.
}
\label{tab:main_and_ablation}
\adjustbox{max totalsize={\textwidth}{0.65\textheight}}{%
  \begin{minipage}{0.45\textwidth}
    \centering
    \begin{tabular}{lcccc}
    \toprule
    \textbf{Method} & \textbf{AUPRC} & \textbf{AUROC} & \textbf{P@100} & \textbf{AUPRC Ratio} \\
    \midrule
    \multicolumn{5}{l}{\textbf{mN (tomato)}}\\
    WGCNA                 & 0.03349 & 0.5080 & 0.11 & 1.0665 \\
    GRNBoost2             & 0.04505 & 0.6534 & 0.00 & 1.4347 \\
    scGPT+head            & 0.03139 & 0.4998 & 0.01 & 0.9999 \\
    scGLUE+head           & 0.03312 & 0.5055 & 0.17 & 1.0550 \\
    SCENIC+               & 0.03160 & 0.5008 & 0.13 & 1.0062 \\
    Pando                 & 0.03160 & 0.5009 & 0.11 & 1.0076 \\
    \cmidrule(lr){1-5}
    RNA-only              & 0.05238 & 0.6550 & 0.16 & 1.6683 \\
    Gene-peak rand.       & 0.06561 & 0.6964 & 0.17 & 2.0897 \\
    \method               & \textbf{0.08458} & \textbf{0.7415} & 0.13 & \textbf{2.6939} \\
    \methodnnpu           & 0.07081 & 0.6862 & \textbf{0.25} & 2.2553 \\
    \midrule
    \multicolumn{5}{l}{\textbf{pN (tomato)}}\\
    WGCNA                 & 0.03307 & 0.5055 & 0.23 & 1.0532 \\
    GRNBoost2             & 0.04435 & 0.6435 & 0.00 & 1.4126 \\
    scGPT+head            & 0.03139 & 0.4998 & 0.01 & 0.9999 \\
    scGLUE+head           & 0.03173 & 0.5016 & 0.07 & 1.0105 \\
    SCENIC+               & 0.03180 & 0.5019 & 0.09 & 1.0134 \\
    Pando                 & 0.03190 & 0.5013 & 0.17 & 1.0144 \\
    \cmidrule(lr){1-5}
    RNA-only              & \textbf{0.08173} & 0.7106 & 0.24 & \textbf{2.6058} \\
    Gene-peak rand.       & 0.06917 & 0.6910 & 0.23 & 2.2053 \\
    \method               & 0.07451 & \textbf{0.7340} & 0.16 & 2.3756 \\
    \methodnnpu           & 0.07090 & 0.6958 & \textbf{0.26} & 2.2605 \\
    \bottomrule
    \end{tabular}
  \end{minipage}\hfill
  \begin{minipage}{0.45\textwidth}
    \centering
    \begin{tabular}{lcccc}
    \toprule
    \textbf{Method} & \textbf{AUPRC} & \textbf{AUROC} & \textbf{P@100} & \textbf{AUPRC Ratio} \\
    \midrule
    \multicolumn{5}{l}{\textbf{PBMC (human)}}\\
    WGCNA                 & 0.00040 & 0.5151 & 0.00 & 1.3174 \\
    GRNBoost2             & 0.00041 & 0.5713 & 0.00 & 1.3677 \\
    scGPT+head            & 0.00046 & 0.5059 & 0.01 & 1.5338 \\
    scGLUE+head           & 0.00034 & 0.5147 & 0.00 & 1.1371 \\
    SCENIC+               & 0.00036 & 0.5059 & 0.01$^{*}$ & 1.1950 \\
    Pando                 & 0.00064 & 0.5071 & \textbf{0.03} & 2.1251 \\
    \cmidrule(lr){1-5}
    RNA-only              & 0.00165 & 0.7782 & 0.00 & 5.4754 \\
    Gene-peak rand.       & 0.00150 & 0.7540 & 0.00 & 4.9777 \\
    \method               & \textbf{0.00244} & \textbf{0.8218} & \textbf{0.03} & \textbf{8.0970} \\
    \methodnnpu           & 0.00135 & 0.7459 & 0.01 & 4.4799 \\
    \midrule
    \multicolumn{5}{l}{\textbf{Mouse brain}}\\
    WGCNA                 & 0.00011 & 0.5074 & 0.00 & 1.2497 \\
    GRNBoost2             & 0.00010 & 0.5295 & 0.00 & 1.1048 \\
    scGPT+head            & 0.00015 & 0.5936 & \textbf{0.01} & 1.7334 \\
    scGLUE+head           & 0.00011 & 0.5302 & 0.00 & 1.2245 \\
    SCENIC+               & 0.00009 & 0.5020 & 0.00 & 1.0188 \\
    Pando                 & 0.00010 & 0.5018 & 0.00 & 1.0874 \\
    \cmidrule(lr){1-5}
    RNA-only              & 0.00009 & 0.5205 & 0.00 & 1.0191 \\
    Gene-peak rand.       & 0.00014 & 0.5715 & 0.00 & 1.5853 \\
    \method               & \textbf{0.00057} & \textbf{0.7263} & \textbf{0.01} & \textbf{6.4543} \\
    \methodnnpu           & 0.00027 & 0.5170 & \textbf{0.01} & 3.0573 \\
    \bottomrule
    \end{tabular}
  \end{minipage}
}
\end{table*}

\begin{figure*}[t]
  \centering
  \adjustbox{max totalsize={\textwidth}{0.72\textheight}}{%
    \begin{minipage}{\textwidth}
      \centering
      \subfloat{
        \includegraphics[width=0.85\linewidth]{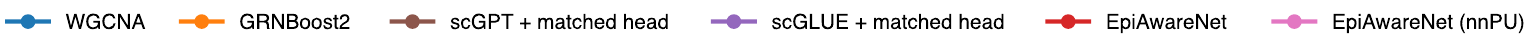}
      }\\[-0.3em]
      \subfloat[PR curve for mN\label{fig:mn-pr}]{
        \includegraphics[width=0.22\linewidth]{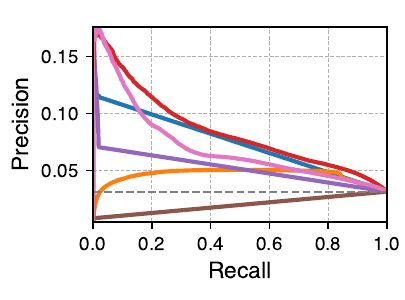}
      }\hfill
      \subfloat[ROC curve for mN\label{fig:mn-roc}]{
        \includegraphics[width=0.22\linewidth]{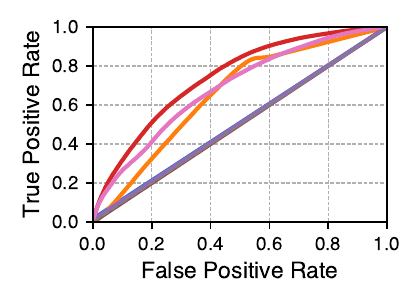}
      }\hfill
      \subfloat[Precision@K for mN\label{fig:mn-patk}]{
        \includegraphics[width=0.22\linewidth]{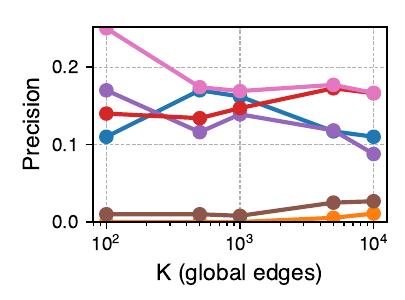}
      }\hfill
      \subfloat[Hit@K for mN\label{fig:mn-hitk}]{
        \includegraphics[width=0.22\linewidth]{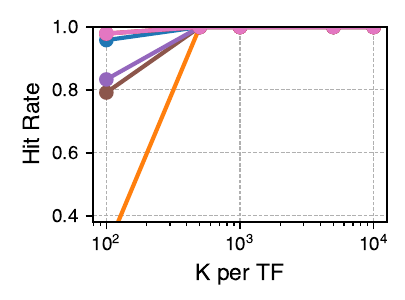}
      }


      \subfloat[PR curve for pN\label{fig:pn-pr}]{
        \includegraphics[width=0.22\linewidth]{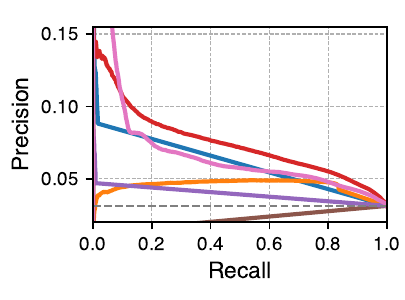}
      }\hfill
      \subfloat[ROC curve for pN\label{fig:pn-roc}]{
        \includegraphics[width=0.22\linewidth]{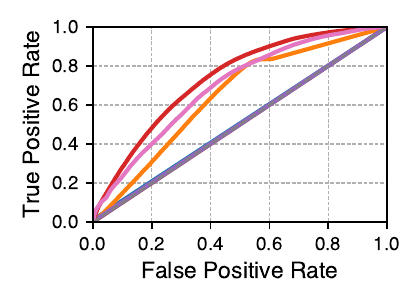}
      }\hfill
      \subfloat[Precision@K for pN\label{fig:pn-patk}]{
        \includegraphics[width=0.22\linewidth]{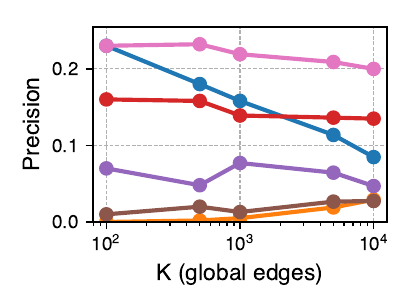}
      }\hfill
      \subfloat[Hit@K for pN\label{fig:pn-hitk}]{
        \includegraphics[width=0.22\linewidth]{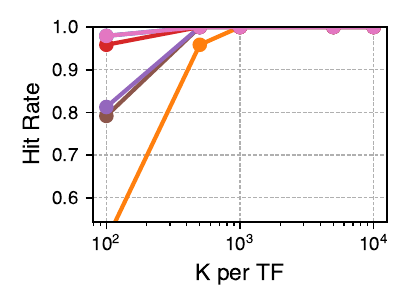}
      }
    \end{minipage}
  }
  \caption{PR, ROC, Precision@K, and Hit@K curves for pN and mN datasets.}
  \label{fig:pn-mn-curves}
\end{figure*}

Table~\ref{tab:main_and_ablation} summarizes overall GRN accuracy across the four datasets.

\paragraph{Classical RNA-only baselines.}
We compare \method with WGCNA and GRNBoost2~\cite{Langfelder2008,Moerman2019}.
On \texttt{mN}/\texttt{pN}, \method substantially improves global ranking: AUPRC increases from 0.04505 / 0.04435 (GRNBoost2) to 0.08458 / 0.07451 (\(+88\%\)/\(+68\%\)), and AUROC improves from 0.6534 / 0.6435 to 0.7415 / 0.7340.
\method also achieves the best AUPRC ratio on tomato (2.6939 on \texttt{mN}; 2.3756 on \texttt{pN}), indicating gains beyond differences in positive prevalence.
When we focus on the very top of the ranked list, the differences look different.
On pN, WGCNA attains higher P@100 (0.23) than \method (0.16), while \methodnnpu achieves the best early precision (P@100 = 0.26).
Overall, \method and \methodnnpu reflect two complementary goals: \method targets \emph{global ranking quality} (AUPRC/AUROC), whereas \methodnnpu aims to improve \emph{very-top retrieval} (P@100).




\paragraph{Deep representation baselines.}
We next compare with scGPT and scGLUE~\cite{Cui2024,Cao2022GLUE} using a matched prediction head.
Relative to scGPT and scGLUE, \method achieves the best AUPRC, AUPRC ratio, and AUROC in every setting (Table~\ref{tab:main_and_ablation}).
For early retrieval, \method ties with Pando on PBMC (P@100 = 0.03) and ties on mouse brain (0.01), while \methodnnpu is strongest on tomato (P@100 = 0.25/0.26 on \texttt{mN}/\texttt{pN}) but with weaker AUPRC/AUROC than \method.
In other words, \methodnnpu improves precision at a fixed small budget, but at the cost of weaker global ranking (smaller AUPRC/AUROC than \method), which is consistent with nnPU encouraging a more conservative scoring function that concentrates mass on a smaller set of high-confidence edges.

\paragraph{Recent paired multi-omic pipelines.}
Table~\ref{tab:main_and_ablation} further compares against SCENIC+ and Pando under the same fixed candidate edge space.
Across all four datasets, \method achieves higher AUPRC, AUPRC ratio, and AUROC than both newer baselines, with especially large AUROC gains over their near-random rankings in this dense candidate space.
These results suggest that adaptive gene--peak aggregation and weakly supervised regulator--target scoring are beneficial beyond heuristic multi-omic pipelines.



\paragraph{Global ranking vs.\ early retrieval.}
Table~\ref{tab:main_and_ablation} summarizes performance with scalar metrics (AUPRC/AUROC and P@100), and Figure~\ref{fig:pn-mn-curves} further provides a diagnostic view across different thresholds by plotting PR/ROC curves and budgeted retrieval curves on tomato datasets. 
From the figure, we can see that \method outperforms baselines from two aspects:

\textbf{Global ranking quality (PR/ROC).}
On mN, the PR and ROC curves show that \method outperforms baselines over a broad range of thresholds, leading to the best AUPRC (0.08458) and AUROC (0.7415).
On pN, \method still achieves the best AUROC (0.7340), although the RNA-only ablation attains a slightly higher scalar AUPRC (0.08173 vs.\ 0.07451), indicating that the benefit of ATAC is dataset-dependent rather than uniformly dominant.
Compared to \method, \methodnnpu typically exhibits slightly weaker PR/ROC behavior, aligning with its lower AUPRC/AUROC in Table~\ref{tab:main_and_ablation}.

\textbf{Early retrieval (Precision@K / Hit@K).}
The Precision@K curves show that, at small budgets, \methodnnpu retrieves higher-precision edge sets on tomato datasets (e.g., P@100 = 0.25/0.26 on mN/pN versus 0.13/0.16 for \method).
The Hit@K curves show how many held-out links are recovered as $K$ increases; here, the two objectives can rank the very top edges differently.

Taken together, Figure~\ref{fig:pn-mn-curves} clarifies the balance between \method and \methodnnpu: \method is optimized for robust global prioritization, while \methodnnpu provides a precision-oriented alternative for budgeted screening.
Additional curves on the other datasets, PBMC and mouse brain, are provided in Appendix Figure~\ref{fig:mouse_brain-pbmc-curves}.

Due to page limit, we defer the computational efficiency analysis in Appendix~\ref{sec:scalability}. 
The time complexity of \method remains low as cell and gene number $N$ and $G$ increases, which demonstrates good empirical \textbf{scalability} of our method.

\subsection{Ablation Studies}
Table~\ref{tab:main_and_ablation} also summarizes the ablation results.
To isolate which components drive the gains of \method{}, we ablate a few key design choices:
(i) \textbf{multiome integration} by removing ATAC evidence (RNA-only),
(ii) \textbf{candidate gene--peak construction} by randomizing gene--peak candidates, and
All variants share the same stage-2 head architecture and optimization recipe, so the differences primarily reflect changes in upstream evidence and candidate construction rather than tuning.

\paragraph{Findings.}
\textbf{Removing ATAC (RNA-only).}


Removing ATAC reduces performance on \texttt{mN}, PBMC, and mouse brain (e.g., AUPRC: 0.08458$\rightarrow$0.05238 on \texttt{mN}; 0.00244$\rightarrow$0.00165 on PBMC; 0.00057$\rightarrow$0.00009 on mouse brain), supporting that accessibility provides complementary evidence for global ranking under severe class imbalance.
On \texttt{pN}, the RNA-only variant attains slightly higher AUPRC (0.08173 vs.\ 0.07451) and competitive P@100 (0.24 vs.\ 0.16), suggesting stronger RNA signal and/or noisier gene--peak evidence in this condition.
Nevertheless, the full model still achieves the best AUROC on \texttt{pN} (0.7340 vs.\ 0.7106 for RNA-only), and adding ATAC improves AUPRC on the other three datasets, in some cases substantially.
We therefore interpret the contribution as improved overall robustness across datasets rather than uniform improvement on every metric for every benchmark.

\textbf{Effect of gene--peak randomization.}
Randomizing gene--peak candidates consistently degrades performance relative to \method (e.g., mN AUPRC 0.08458 to 0.06561), with similar decreases in AUROC.
This supports our candidate construction as a key ingredient for making cross-modal (RNA+ATAC) reasoning helpful.


\subsection{Robustness Analysis}
\label{sec:robustness}

\begin{figure}[t]
  \centering
  \subfloat[AUPRC\label{fig:noise-auprc}]{
    \includegraphics[width=0.45\linewidth]{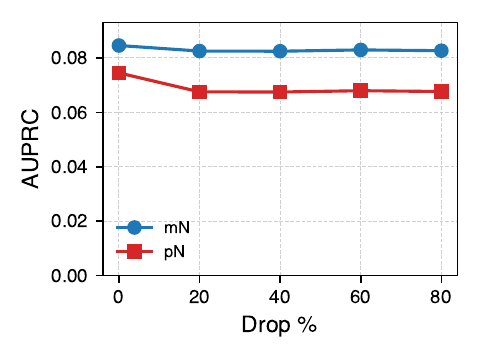}
  }\hfill
  \subfloat[AUROC\label{fig:noise-auroc}]{
    \includegraphics[width=0.45\linewidth]{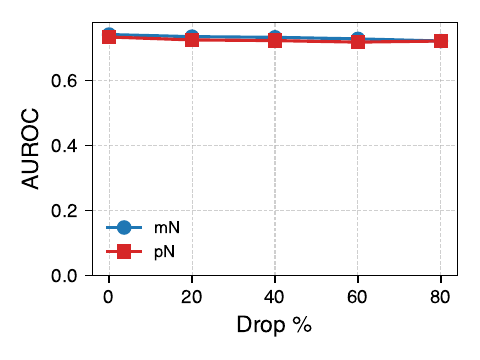}
  }

  \caption{Robustness to noise in the bulk-derived GRN prior. We report AUPRC (left) and AUROC (right) as the prior is progressively perturbed on the pN and mN datasets.}

  \label{fig:noise-robustness}
  \vspace{-1.5em}
\end{figure}

Figure~\ref{fig:noise-robustness} evaluates Stage~2 robustness to noise in the bulk-derived GRN prior by progressively dropping a fraction of prior edges and reporting performance on pN and mN. As the drop rate increases from 0\% to 80\%, AUPRC decreases initially (around 20\% drop) and then largely plateaus, indicating limited sensitivity once a minimal amount of weak supervision remains. In contrast, AUROC declines more smoothly and approximately monotonically as the drop rate increases, but the overall degradation remains modest on both datasets. These trends suggest that the learned multi-omic representations contribute signal beyond simply reproducing the Stage~2 prior, while the bulk prior provides directed weak supervision and a lightweight regularizing guide that improves calibration and ranking against harder negatives. As an additional check, Appendix~\ref{app:additional_robustness_transfer} reports cross-database evaluation: PBMC models trained with DoRothEA remain enriched on TRRUST-human links, and mouse-brain models trained with TRRUST remain enriched on DoRothEA-mouse links.

\subsection{Case Study}

\begin{figure}[t]
  \centering
  \includegraphics[width=0.8\linewidth]{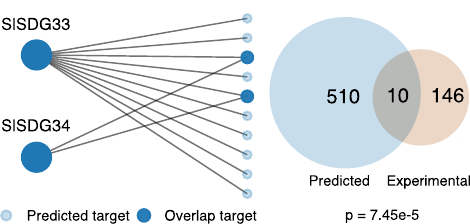}
  \caption{Case study on tomato pN: For unseen regulators SlSDG33/SlSDG34, \method predictions show statistically significant overlaps with experiments from literature.}
  \label{fig:case-study}
  \vspace{-1.5em}
\end{figure}

While the above evaluations focus on regulators observed during training, we further assess \method through a case study on previously unseen regulators.
Specifically, we analyze two chromatin regulators, SlSDG33 and SlSDG34, which were not included in the training set. Both regulators are histone methyltransferases implicated in nitrogen response pathways in tomato roots.
In the pN dataset, we consider the top 146 ranked target-gene predictions for SlSDG33 and SlSDG34. A prior experimental study identified 510 genes regulated by these two chromatin regulators during nitrogen response using molecular assays.
Despite the incomplete and noisy nature of regulatory annotations, the computationally inferred targets show a statistically significant overlap of 10 genes with the experimentally determined set (hypergeometric enrichment p-value = $7.45\times 10^{-5}$). As illustrated in Figure~\ref{fig:case-study}, overlapping genes are highlighted within the predicted subnetwork.
This result demonstrates that \method can generalize to unseen regulators and recover biologically meaningful targets, with predictions supported by independent in planta experimental evidence. Detailed rank-level enrichment statistics are provided in Appendix~\ref{app:case_rank_enrichment}.

\section{Conclusion}
\label{sec:conclusion}
We presented \method, a prior-guided multi-omic Transformer framework for GRN inference from paired scRNA-seq and scATAC-seq data.
\method combines (i) candidate-constrained, gene-centric cross-attention to integrate chromatin accessibility evidence without relying on hard-coded peak-to-gene links, and (ii) a weakly supervised Stage~2 that refines regulator--target predictions using bulk-derived regulatory links under label scarcity.
Across multiple paired multiome datasets, our results show that incorporating lightweight priors together with adaptive cross-modal representation learning improves GRN reconstruction and yields biologically plausible regulatory hypotheses.
Looking ahead, promising directions include enriching candidate construction to better capture distal regulation and extending inference to context-specific GRNs across cell types or conditions while preserving efficiency.

\section{Limitations and Ethical Considerations}
\label{sec:limitations_ethics}

\paragraph{Limitations.}
\method trades flexibility for efficiency and stability in several places.
Stage~1 aggregates chromatin evidence via gene-centric cross-attention over precomputed candidate peak sets; while this candidate constraint reduces computation, it can miss distal regulatory elements when candidate construction is incomplete or overly local, and it may inherit biases from how candidate neighborhoods are defined.
Stage~2 uses bulk-derived regulator--target links as weak supervision (noisy positives) in a positive--unlabeled setting; if the prior is biased toward well-studied pathways, tissues, or species, or lacks condition-specific regulation, it may be misaligned with the true single-cell context and affect edge ranking.
Accordingly, our benchmark evaluates weakly supervised generalization to unseen held-out edges under a prior-informed regime, rather than prior-free de novo GRN discovery.
Evaluation also relies on imperfect gold references that are incomplete and may contain false negatives, so global ranking metrics (e.g., AUPRC/AUROC) can understate biologically valid discoveries and should be interpreted together with complementary evidence (e.g., early-retrieval metrics and orthogonal analyses).

\paragraph{Ethical Considerations.}
\method is intended to support scientific discovery by prioritizing candidate regulatory hypotheses from paired single-cell measurements; inferred GRNs should not be treated as validated causal mechanisms without follow-up evidence.
Because the model leverages prior knowledge (e.g., bulk-derived regulatory links) as weak supervision, it may reflect historical biases in available resources and preferentially rank well-characterized interactions.
We therefore recommend transparent reporting of prior sources and candidate construction procedures when drawing biological conclusions.
When applying the method to human datasets, practitioners should comply with dataset governance and institutional policies, use appropriate de-identification and access control, and avoid making decisions about identifiable individuals based on inferred networks.

\section*{GenAI Disclosure}
The authors confirm that the majority of the text in this paper was written by the authors themselves. We used GPT-5.2 to assist with brainstorming ideas, editing for clarity, and improving language fluency, and Copilot to assist with coding. Any content generated by AI tools has been reviewed and edited by the authors, and we take full responsibility for the final content of the paper.

\section*{Acknowledgement}
This research is supported in part by an Ag-Eng seed grant to Y.L. and J.G. from Purdue university and the US National Science Foundation under grant NSF IIS-2141037. Any opinions, findings, and conclusions or recommendations expressed in this material are those of the author(s) and do not necessarily reflect the views of the National Science Foundation.


\bibliographystyle{ACM-Reference-Format}
\bibliography{mybib}

\appendix

\begingroup
\setlength{\textfloatsep}{6pt plus 1pt minus 1pt}
\setlength{\floatsep}{6pt plus 1pt minus 1pt}
\setlength{\intextsep}{6pt plus 1pt minus 1pt}

\section{Data Preparation}\label{app:data_prep}
For scRNA-seq, we start from cell-by-gene UMI matrices; for scATAC-seq, we start from cell-by-peak fragment matrices after peak calling and read assignment. Paired multiome assays share cell barcodes by construction; otherwise, we intersect and reorder barcodes to align the two modalities. Standard QC removes low-quality cells and low-coverage genes/peaks. The default dataloader applies $\log(1+x)$ to both modalities. For each gene $g$, we construct a candidate peak set $\mathcal{C}_g\subseteq\mathcal{P}$ by selecting peaks within a $\pm5$kb window around the gene, ranking candidates by genomic distance, and truncating to a fixed maximum width for batching. Bulk-derived regulator--target priors are mapped into the model gene vocabulary; matched edges are used as noisy positives in Stage~2 and all other regulator--target pairs remain unlabeled.
\section{Implementation Details}\label{app:impl_details}

\subsection{EpiAwareNet implementation}\label{app:attention}
Stage~1 uses the same backbone across datasets: $L=6$ Transformer layers, hidden dimension $d=256$, 8 heads, dropout 0.1, sparse gene--gene neighborhoods, and gene--peak cross-attention restricted to $\mathcal{C}_g$ with optional Top-$K$ routing (default $K=8$). Candidate lists are capped at 50 peaks per gene and padded/truncated for batching. We pretrain with 15\% random gene masking and an NB-style decoder over preprocessed expression values. Stage~1 optimization uses AdamW with learning rate $10^{-4}$, weight decay $10^{-2}$, batch size 64, seed 42, and early stopping by validation reconstruction loss. Stage~2 freezes the backbone and trains a one-hidden-layer MLP head (hidden size 256, ReLU, dropout 0.1) on concatenated regulator--target embeddings. Unless otherwise stated, Stage~2 uses BCE with bulk-derived \texttt{train} links as positives, a 10:1 sampled unlabeled-to-positive ratio, AdamW with learning rate $3\times10^{-4}$ and weight decay $10^{-2}$, 20 epochs, and no \texttt{val} edges for checkpoint selection or tuning.

\subsection{Baseline implementation}\label{app:baselines_impl}
GRNBoost2 and WGCNA are run on the same normalized RNA matrix as \method; GRNBoost2 uses default per-target feature-importance scores, while WGCNA converts the undirected TOM similarity into regulator-anchored directed scores by ranking each regulator's top targets. For scGPT, we train on the same normalized scRNA-seq matrix, extract final-layer gene token embeddings, and fit the same Stage~2 head using the same regulator list and \texttt{train} prior edges. For scGLUE, we fit paired RNA/ATAC embeddings using the same gene--candidate-peak mapping as structural guidance and again use the matched Stage~2 head. SCENIC+ and Pando are run on paired RNA/ATAC inputs and evaluated on the same candidate edge space $\mathcal{T}\times\mathcal{G}$; unreturned edges receive score 0 before computing AUPRC/AUROC.

\section{Complexity Analysis}\label{app:complexity}
Let $d$ be the hidden dimension, $L$ the number of Transformer layers, and $\overline{|\mathcal{C}|}$ the average candidate-set size. Full gene--gene attention costs $O(NG^2d)$ per layer; restricting each gene to $k_{\mathrm{NN}}\ll G$ neighbors gives $O(NGk_{\mathrm{NN}}d)$. Full gene--peak attention would cost $O(NGPd)$, whereas candidate-constrained attention costs $O(N\sum_g |\mathcal{C}_g|d)\approx O(NG\overline{|\mathcal{C}|}d)$, with optional Top-$K$ routing further limiting the active peak set. Thus the dominant backbone cost scales with sparse neighborhoods and capped gene-specific peak candidates rather than dense cell--cell, gene--gene, or gene--peak pairwise graphs.

\section{Additional Results}

\begin{figure*}[t]
  \centering
  \adjustbox{max totalsize={\textwidth}{0.72\textheight}}{%
    \begin{minipage}{\textwidth}
      \centering
      \subfloat{
        \includegraphics[width=0.75\linewidth]{figures/curves/legend.pdf}
      }\\[-0.3em]
      \subfloat[PR curve for PBMC\label{fig:pbmc-pr}]{
        \includegraphics[width=0.22\linewidth]{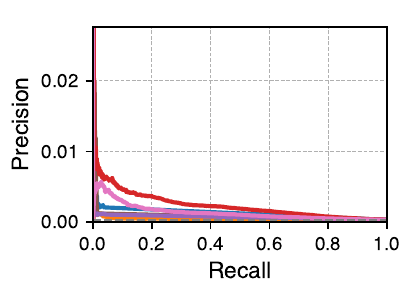}
      }\hfill
      \subfloat[ROC curve for PBMC\label{fig:pbmc-roc}]{
        \includegraphics[width=0.22\linewidth]{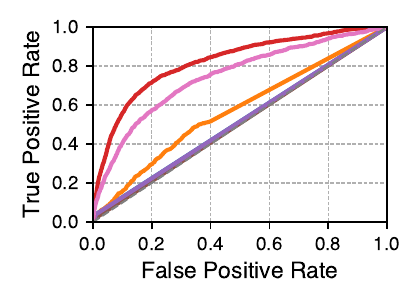}
      }\hfill
      \subfloat[Precision@K for PBMC\label{fig:pbmc-patk}]{
        \includegraphics[width=0.22\linewidth]{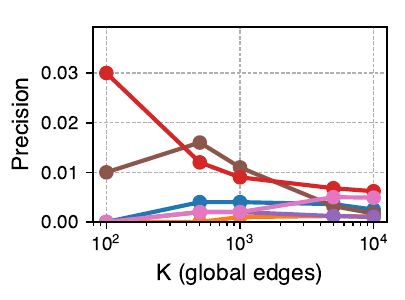}
      }\hfill
      \subfloat[Hit@K for PBMC\label{fig:pbmc-hitk}]{
        \includegraphics[width=0.22\linewidth]{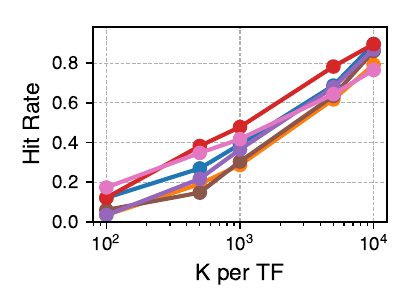}
      }


      \subfloat[PR curve for Mouse Brain\label{fig:mouse_brain-pr}]{
        \includegraphics[width=0.22\linewidth]{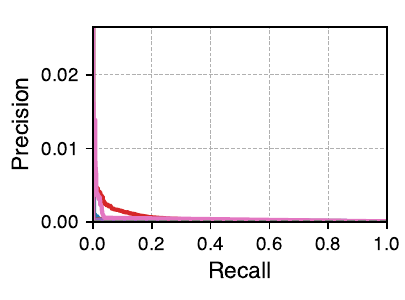}
      }\hfill
      \subfloat[ROC curve for Mouse Brain\label{fig:mouse_brain-roc}]{
        \includegraphics[width=0.22\linewidth]{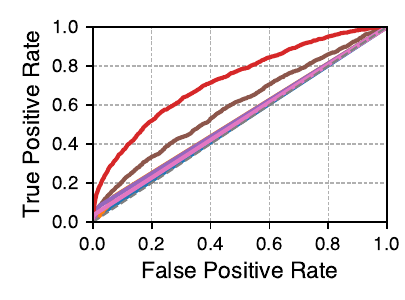}
      }\hfill
      \subfloat[Precision@K for Mouse Brain\label{fig:mouse_brain-patk}]{
        \includegraphics[width=0.22\linewidth]{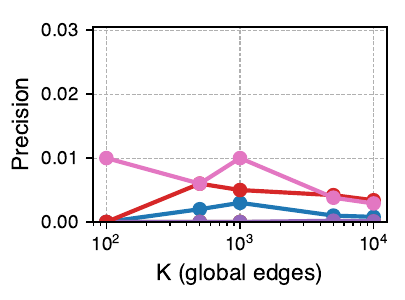}
      }\hfill
      \subfloat[Hit@K for Mouse Brain\label{fig:mouse_brain-hitk}]{
        \includegraphics[width=0.22\linewidth]{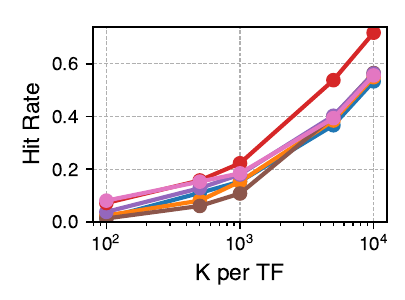}
      }
    \end{minipage}
  }
  \caption{PR, ROC, Precision@K, and Hit@K curves for PBMC and Mouse Brain datasets.}
  \label{fig:mouse_brain-pbmc-curves}
\end{figure*}

\subsection{Additional PR/ROC and top-$K$ behavior}\label{app:additional_curves}
On PBMC, \method achieves stronger global ranking (AUPRC 0.00244 and AUROC 0.8218) than the strongest external baselines (AUPRC 0.00046 for scGPT; AUROC 0.5713 for GRNBoost2), while scGPT is better at the very top of the list (P@1000 0.011 vs. 0.002 for \method). On mouse brain, \method also improves global ranking (AUPRC 0.00057 and AUROC 0.7263) relative to the strongest external baselines (AUPRC 0.00015 for scGPT; AUROC 0.5936 for scGPT). These trends indicate that global PR/ROC ranking and very small-budget retrieval can differ under extreme class imbalance.

\begin{table}[t]
\centering
\small
\setlength{\tabcolsep}{6pt}
\caption{$p$-values for GRNs. Smaller is better.}
\label{tab:pvalues}
\begin{tabular}{lcccc}
\toprule
\textbf{Method} & \textbf{mN} & \textbf{pN} & \textbf{PBMC} & \textbf{Mouse Brain} \\
\midrule
WGCNA        & <0.01 & <0.01 & <0.01 & <0.01 \\
GRNBoost2    & 0.0523    & <0.01    &  <0.01 & <0.01 \\
scGPT        & 0.3980 & 0.0796 & <0.01 & <0.01 \\
scGLUE       & <0.01 & <0.01 & <0.01 & <0.01 \\
EpiAwareNet  & <0.01 & <0.01 & <0.01 & <0.01 \\
\bottomrule
\end{tabular}
\end{table}
\subsection{GRN-level $p$-value}\label{app:pvalue}
For a predicted directed edge set $\hat{E}$, we report the overlap $S(\hat{E})=|\hat{E}\cap E_{\mathrm{val}}|$ with held-out links and compare it to degree-preserving random networks $\{E^{(r)}\}_{r=1}^R$ that keep each regulator's out-degree fixed while randomly reassigning targets. The one-sided empirical value is $p=(1+|\{r:S(E^{(r)})\ge S(\hat{E})\}|)/(R+1)$. Table~\ref{tab:pvalues} shows that \method is significant across all datasets, whereas some baselines are weaker on tomato.

\subsection{Rank-level enrichment for unseen regulators}\label{app:case_rank_enrichment}
For SlSDG33/SlSDG34, the independent validated reference contains 510 genes and the case-study ranking universe contains 34{,}075 genes. Among the top 146 ranked predictions, 10 overlap the validated set, compared with 2.19 expected by random ranking. This gives 4.58$\times$ enrichment (hypergeometric $p=7.45\times10^{-5}$); validated targets occur at ranks \{5, 19, 38, 57, 64, 75, 102, 104, 118, 120\}, showing concentration near the top rather than random scattering.

\begin{table}[t]
\centering
\setlength{\tabcolsep}{5pt}
\renewcommand{\arraystretch}{1.03}
\scriptsize
\caption{Cumulative enrichment of independently validated SlSDG33/SlSDG34 targets among top-ranked \method predictions.}
\label{tab:case_rank_enrichment}
\begin{tabular}{lccc}
\toprule
\textbf{Cutoff} & \textbf{Observed} & \textbf{Expected} & \textbf{Enrichment} \\
\midrule
Top-5 & 1 & 0.07 & 13.3$\times$ \\
Top-50 & 3 & 0.75 & 4.0$\times$ \\
Top-100 & 6 & 1.50 & 4.0$\times$ \\
Top-146 & 10 & 2.19 & 4.58$\times$ \\
\bottomrule
\end{tabular}
\end{table}

\subsection{Additional robustness, transfer, and scalability}\label{app:additional_robustness_transfer}
Cross-database tests remain enriched over random ranking (Table~\ref{tab:cross_database_transfer}). Broader gene--peak windows improve mammalian AUROC, with $\pm100$kb giving mouse/PBMC AUROC 0.8002/0.8708; $\pm50$kb gives the best mouse AUPRC (0.001018), and $\pm100$kb gives the best PBMC AUPRC (0.002933). Three additional seeds \{43,44,45\} show low variance on PBMC ($0.002475\pm4.55{\times}10^{-5}$ AUPRC; $0.8232\pm7.57{\times}10^{-4}$ AUROC) and mouse brain ($0.000563\pm1.0{\times}10^{-6}$ AUPRC; $0.7317\pm1.63{\times}10^{-3}$ AUROC).

\begin{table}[t]
\centering
\setlength{\tabcolsep}{3pt}
\renewcommand{\arraystretch}{1.03}
\scriptsize
\caption{Cross-database evaluation using independent regulatory links as test gold standards.}
\label{tab:cross_database_transfer}
\adjustbox{max width=\linewidth}{%
\begin{tabular}{l l r c c}
\toprule
\textbf{Evaluation} & \textbf{Gold} & \textbf{\#Edges} & \textbf{AUROC} & \textbf{AUPRC ratio} \\
\midrule
PBMC (DoRothEA-trained) & TRRUST-human & 5{,}374 & 0.6225 & 1.81$\times$ \\
Mouse brain (TRRUST-trained) & DoRothEA-mouse & 4{,}564 & 0.6100 & 2.02$\times$ \\
\bottomrule
\end{tabular}
}
\end{table}

\begin{table}[t]
\centering
\setlength{\tabcolsep}{3pt}
\renewcommand{\arraystretch}{1.03}
\scriptsize
\caption{Sensitivity to the gene--peak candidate window on mammalian datasets. Best value within each metric is in \textbf{bold}.}
\label{tab:window_sensitivity}
\adjustbox{max width=\linewidth}{%
\begin{tabular}{l c c c c}
\toprule
\textbf{Window} & \textbf{Mouse AUPRC} & \textbf{Mouse AUROC} & \textbf{PBMC AUPRC} & \textbf{PBMC AUROC} \\
\midrule
$\pm 5$kb & 0.000570 & 0.7263 & 0.002440 & 0.8218 \\
$\pm 50$kb & \textbf{0.001018} & 0.7662 & 0.002864 & 0.8673 \\
$\pm 100$kb & 0.000779 & \textbf{0.8002} & \textbf{0.002933} & \textbf{0.8708} \\
\bottomrule
\end{tabular}
}
\end{table}

\begin{table}[t]
\centering
\setlength{\tabcolsep}{4pt}
\renewcommand{\arraystretch}{1.03}
\scriptsize
\caption{Seed stability over three additional seeds \{43,44,45\}. Entries are mean $\pm$ standard deviation.}
\label{tab:seed_stability}
\begin{tabular}{l c c}
\toprule
\textbf{Dataset} & \textbf{AUPRC} & \textbf{AUROC} \\
\midrule
PBMC & $0.002475 \pm 4.55{\times}10^{-5}$ & $0.8232 \pm 7.57{\times}10^{-4}$ \\
Mouse brain & $0.000563 \pm 1.0{\times}10^{-6}$ & $0.7317 \pm 1.63{\times}10^{-3}$ \\
\bottomrule
\end{tabular}
\end{table}

\subsection{Scalability}\label{sec:scalability}
\begin{figure}[H]
  \centering
  \subfloat[Runtime vs.\ number of cells $N$\label{fig:scalability-time-vs-n}]{
    \includegraphics[width=0.47\linewidth]{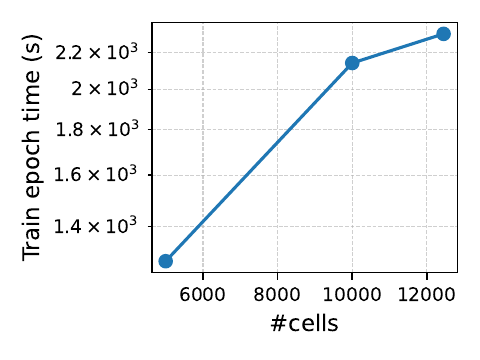}
  }\hfill
  \subfloat[Runtime vs.\ number of genes $G$\label{fig:scalability-time-vs-g}]{
    \includegraphics[width=0.47\linewidth]{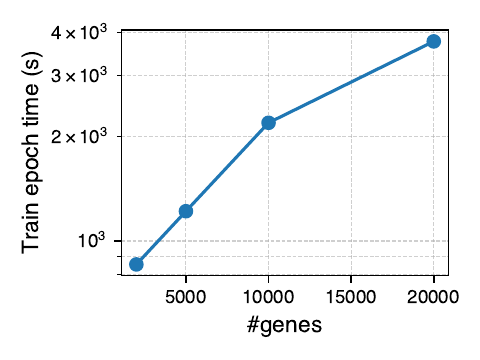}
  }
  \caption{Scalability of \method on tomato pN.}
  \label{fig:scalability}
\end{figure}
On tomato pN, end-to-end runtime increases smoothly and approximately linearly as the number of cells $N$ or genes $G$ increases under fixed settings (Figure~\ref{fig:scalability}), without the sharp blow-up expected from naive dense pairwise modeling such as $O(N^2)$ over cells or $O(G^2)$ over genes.
This trend is consistent with the design of \method: computation is dominated by processing observed cells and routing over capped gene-specific peak candidate sets, rather than materializing dense all-gene or all-peak interactions.
We therefore use these curves as an empirical sanity check that the sparse attention design remains practical when scaling the input matrix along either axis.
In the cell-scaling experiment, we subsample cells while keeping the gene and candidate-peak configuration fixed, so the trend mainly reflects the cost of processing additional single-cell observations.
In the gene-scaling experiment, we vary the number of retained genes while keeping the cell subset fixed, so the trend reflects both additional gene tokens and their associated candidate peak neighborhoods.
The smooth increase in both settings supports the intended sparse-computation regime: the runtime is governed by the number of observed cells and capped local candidate sets, rather than by constructing dense pairwise relationships over all cells, genes, or peaks.
\endgroup
\end{document}